\documentclass[a4paper,12pt]{article}

\usepackage{graphicx}
\usepackage{booktabs}
\usepackage{hyperref}
\usepackage{graphicx}
\usepackage{pdfsync}
\usepackage{amsmath}
\usepackage{amssymb}
\usepackage{mathrsfs}
\usepackage{color}

\newcommand{\inputspace}{\ensuremath{\mathcal{X}}}
\newcommand{\outputspace}{\ensuremath{\mathcal{Y}}} 
\newcommand{\setoflabels}{\ensuremath{\mathcal{L}}} 
\newcommand{\argmin}{\operatornamewithlimits{argmin}}

\def\vec#1{\mathchoice{\mbox{\boldmath$\displaystyle#1$}}
{\mbox{\boldmath$\textstyle#1$}}
{\mbox{\boldmath$\scriptstyle#1$}}
{\mbox{\boldmath$\scriptscriptstyle#1$}}}
\begin{document}

\title{{\bf Rectifying Classifier Chains for\\ Multi-Label Classification}\thanks{Extended version of:
R.~Senge, J.\ del Coz, E.~H\"ullermeier. Rectifying Classifier Chains for Multi-Label Classification. Proceedings Workshop LWA--2013, Lernen--Wissensentdeckung--Adaptivität,151--158, Bamberg, Germany, 2013.
%This research has been supported by the Germany Research Foundation (DFG) and the Spanish Ministerio de Econom\'ia y Competitividad under grant TIN2011-23558.
}}


%\titlerunning{Rectifying Classifier Chains}        % if too long for running head

%\email{\{senge,eyke\}@informatik.uni-marburg.de \\ juanjo@aic.uniovi.es}

%\author{Robin Senge$^1$, Juan Jos\'e del Coz$^2$ and Eyke H\"ullermeier$^1$}

%\institute{$^1$ Department of Mathematics and Computer Science, Marburg University, 35032 Marburg, Germany \and $^2$ Artificial Intelligence Center, University of Oviedo at Gij\'on, Campus de Viesques, 33204  Gij\'on, Spain}

%\authorrunning{Senge, Coz, H\"ullermeier} % if too long for running head

\author{Robin Senge$^1$, Juan Jos\'e del Coz$^2$ and Eyke H\"ullermeier$^1$\\
	$^1$Heinz Nixdorf Institute and Department of Computer Science\\
Paderborn University, Germany  \\
$^2$ Artificial Intelligence Center, University of Oviedo at Gij\'on,\\
 Campus de Viesques, 33204  Gij\'on, Spain
	}
	
\date{}


\maketitle

\begin{abstract}
Classifier chains have recently been proposed as an appealing method for tackling the multi-label classification task. In addition to several empirical studies showing its state-of-the-art performance, especially when being used in its ensemble variant, there are also some first results on theoretical properties of classifier chains. Continuing along this line, we analyze the influence of a potential pitfall of the learning process, namely the discrepancy between the feature spaces used in training and testing: While true class labels are used as supplementary attributes for training the binary models along the chain, the same models need to rely on estimations of these labels at prediction time. We elucidate under which circumstances the attribute noise thus created can affect the overall prediction performance. As a result of our findings, we propose two modifications of classifier chains that are meant to overcome this problem. Experimentally, we show that our variants are indeed able to produce better results in cases where the original chaining process is likely to fail.\\[2mm]
\end{abstract}

\section{Introduction}
\label{sec:intro}

%:intro
Multi-label classification (MLC) has attracted increasing attention in the machine learning community during the past few years. Apart from being interesting theoretically, this is largely due to its practical relevance in  many domains, including text classification, media content tagging and bioinformatics, just to mention a few. The goal in MLC is to induce a model that assigns a \emph{subset} of labels to each example, rather than a single one as in multi-class classification. For instance, in a news website, a multi-label classifier can automatically attach several labels---usually called tags in this context---to every article; the tags can be helpful for searching related news or for briefly informing users about their content. 

%:old methods
%Many of the first approaches to MLC were aimed at adapting binary or multi-class learners to handle multi-label data, including decision trees \cite{ClareML-DecisionTrees}, instance-based algorithms \cite{ZhangML-KNN}, neural networks \cite{ZhangML-NN}, support vector machines \cite{WestonMulti-Label}, naive Bayes \cite{McCallumBayesML}, conditional random fields \cite{Ghamrawi} and boosting \cite{SchapireBoostexter}. One may argue, however, that the construction of good multi-label classifiers requires dedicated methods able to exploit the particularities of multi-label data. 

%:label dependence
Current research on MLC is largely driven by the idea that optimal predictive performance cannot be achieved without modeling and exploiting \emph{statistical dependencies} between labels. %In other words: When it is time to predict wether the one label is relevant or not, the usual object descriptions should be extended by informations about the relevance of other labels. 
Roughly speaking, if the relevance of one label may depend on the relevance of other labels, i.e., if their relevance is not statistically independent, then labels should be predicted \emph{simultaneously} and not \emph{separately}. 
This is the main argument against simple \emph{decomposition techniques} such as binary relevance (BR) learning, which splits the original multi-label task into several independent binary classification problems, one for each label.
% under the assumption on label independence, that is, ignoring any information about the interencies between the labels and only using the input variables.  

%:some methods
%However, modeling label dependence is a quite difficult task. Indeed, 
Until now, several methods for capturing label dependence have been proposed in the literature. They can be categorized according to two major properties:
\begin{itemize}
\item[(i)] the size of the subsets of labels for which dependencies are modeled, and 
\item[(ii)] the type of label dependence they seek to capture. 
\end{itemize}
Looking at the first property, there are methods that only consider pairwise relations between labels \cite{WestonMulti-Label,FurnkranzLabelRanking,SchapireBoostexter,ZhangML-NN} and approaches that take into account correlations among larger label subsets \cite{ReadPrunedSets,ClassifierChainsML,TsoumakasRAKEL}; the latter include those that consider the influence of all labels simultaneously \cite{EykeIBLR,GodboleDiscriminative,elenaAID}. Regarding the second criterion, it has been proposed to distinguish between the modeling of \emph{conditional} and \emph{unconditional label dependence}  \cite{EykeBayes,LabelDependenceML}, depending on whether the dependence is conditioned on an instance \cite{EykeBayes,elenaAID,ClassifierChainsML,TsoumakasMedidasError} or describing a kind of global correlation in the label space  \cite{EykeIBLR,GodboleDiscriminative,ZhangML-NN}.

%:focus on classifier chains
In this paper, we focus on a method called \emph{classifier chains} (CC) \cite{ClassifierChainsML}. This method enjoys great popularity, even though it has been introduced only lately. As its name suggests, CC selects an order on the label set---a \textit{chain} of labels---and trains a binary classifier for each label in this order. The difference with respect to BR is that the feature space used to induce each classifier is extended by the previous labels in the chain. These labels are treated as additional attributes, with the goal to model conditional dependence between a label and its predecessors. CC performs particularly well when being used in an ensemble framework, usually denoted as \emph{ensemble of classifier chains} (ECC), which reduces the influence of the label order. 

% move to cc section? 
%Since ECC, randomizes the label order for each sub-model, it eliminates the potentially bad influence of just one awkward label order, which deteriorates the performance of a single CC model. Another important advantage over more complex methods like \emph{label powerset} \cite{TsoumakasOverview} is the moderate computational complexity. CC and ECC maintain a computational complexity of the same order as that of BR. This is sometimes crucial, as many multi-label problems involve large numbers of examples and labels.

%:finding
Our study aims at gaining a deeper understanding of CC's learning process. More specifically, we address an issue that, despite having been noticed \cite{LabelDependenceML}, has not been picked out as an important theme so far: Since information about preceding labels is only available for training, this information has to be replaced by estimations (coming from the corresponding classifiers) at prediction time. As a result, CC has to deal with a specific type of attribute noise: While a classifier is learnt on ``clean'' training data, including the true values of preceding labels, it is applied on ``noisy'' test data, in which true labels are replaced by possibly incorrect predictions. Obviously, this type of noise may affect the performance of each classifier in the chain. More importantly, since each classifier relies on its predecessors, a single false prediction might be propagated and possibly even multiplied along the whole chain. 

%Besides the influence of the label order, there is another critical factor that can also have an impact on the final performance of the whole multi-label classifier. As it will be explained in more detail later, if one gets granular to the CC method, it gets obvious, that it uses a different feature space in the test phase than it was using in the training phase. The true label information, which was used during training is not available in test phase. It has to be replaced by predicted labels, which are of cause only estimations. This circumstance introduces attribute noise in the testing, which was not present in training. Moreover, the own structure of the CC multi-label classifier can make this problem even worse. In the testing phase every classifier depends on the outputs of all previous classifiers in the chain. When the number of labels is large, this \emph{testing noise} can produce some prediction errors that propagate through the chain, especially deteriorating the performance of binary classifiers placed in the last positions. 

%:contributions
The contribution of this paper is twofold. First, we analyze the above problem of ``error propagation'' in classifier chains in more detail. Using both synthetic and real data sets, we design experiments in order to reveal those factors that influence the effect of error propagation in CC. Second, we propose and evaluate modifications of the original CC method that are intended to overcome this problem.

%\textcolor{red}{They main ideas are: i) to eliminate the testing noise by using the same feature space in both, training and testing, and 2) to reduce the size of the chains}. SOMETHING ABOUT THE EXPERIMENTAL RESULTS.

%:paper outline
The rest of the paper is organized as follows. After a brief discussion of related work, we introduce the setting of MLC more formally in Section 3, and then explain the classifier chains method in Section~\ref{sec:CC}. Section~\ref{sec:IID} is devoted to a deeper discussion of the aforementioned pitfalls of CC, along with some first experiments for illustration purposes.\footnote{This section is partly based on \cite{Senge_GfKl_2012}} In Section~\ref{sec:NS}, we introduce modifications of CC and propose a method called \emph{nested stacking}. An empirical study, in which we experimentally compare this method with the original CC approach, is presented in Section 6. The paper ends with a couple of concluding remarks in Section~\ref{sec:conclusions}.

\section{Related Work}

While we are not aware of directly related work in the field of multi-label classification, it is worth to have a look at other types of applications, which, in one way or the other, have to deal with problems caused by the propagation/multiplication of prediction errors. In fact, many methods in which predictions are made in a sequential way are immediately prone to this kind of problem.

\emph{Sequence labeling}, for instance, involves the assignment of a categorical label to each element of a sequence of observed values. A typical example is part of speech tagging: Given a sentence (or even a whole document) as an input, the task is to assign a part of speech to each individual word. Obviously, there is a strong dependency between the labels in a given sequence. Therefore, to make an optimal prediction of the label for a specific word, it is important to take the context of this label into consideration, i.e., the (predicted) labels of nearby words. To this end, quite a number of \emph{structured learning algorithms} have been developed and applied to this task \cite{nguyen07}; examples of such algorithms include hidden Markov models, conditional random fields, as well as methods such as SEARN \cite{searn} and HC-Search \cite{doppa2,doppa}, which combine search (in the output space) and learning.

A specific type of sequence labeling is \emph{sequential partitioning}, a sequential classification task for which longer runs of the same label are encountered \cite{Cohen_05}. Here, instances have a single binary label (like in binary classification). However, the set of instances to be classified at prediction time is not drawn independently; instead, they obey a natural order. As an example, consider the task to identify the signature part of an email. An instance then refers to a line of text, and each line has to be classified as being part of the signature or not. The natural order of the lines is given by the structure of the email. To tackle this problem, the authors in \cite{Cohen_05} propose a specific type of \emph{stacking} approach that bears some resemblance to our method of nested stacking (cf.\ Section~\ref{sec:NS}).

Yet another direction is sequential decision making problems such as planning and reinforcement learning, where the goal is to predict an optimal \emph{sequence of actions}. The problem of error propagation has been noticed and specifically well studied in the field of \emph{imitation learning}. In applications like mobile robot navigation and electronic games, imitation learning aims at imitating an experts policy which comprises an optimal selection of sequential actions. By executing actions, the expert and the imitating machine move from one state to another. Erroneously choosing the wrong action then requires a dynamic (state-dependent) recovery policy, which cannot be achieved by simply imitating the faultless expert policy in this situation. In fact, the erroneous action can lead to a higher probability of subsequent errors \cite{Ross_10}.

%Despite the fact, that the setting is different to multi-label classification, in order to classify an instance, information about predictions of nearby instances may help but potentially also mislead.

Finally, we also mention that problems of this kind are of course not limited to the case of categorical predictions but likewise apply to the prediction of real-valued targets, for example, in time series forecasting or in audio and speech signal processing. 
However, since these applications are quite remote from multi-label classification, or at least less connected than those we discussed above, we refrain from a more detailed discussion here.

\section{Multi-Label Classification}\label{sec:ML}

%:intro iid
%The goal of this section is to analyze some properties of CC. First, we will formally review the problem formulation of multi-label classification and introduce CC, as it is one prominent method, which tries to solve this problem. This also including the notation employed throughout the paper. Then, we discuss an important source of "{}noise"{}, which emerges from the own learning process of CC and which arguably violates one essential and common assumption in machine learning, namely the \emph{iid assumption}. We illustrate the effects of this violation using synthetic and benchmark real-world data sets. 

%:ML
Let $\setoflabels=\{\lambda_1, \lambda_2, \ldots, \lambda_m\}$ be a finite and non-empty set of class labels, and let $\inputspace$ be an instance space. We consider an MLC task with a training set 
$$
S = \big\{(\vec{x}_1, \vec{y}_1), \ldots , (\vec{x}_n, \vec{y}_n) \big\} \, 
$$
generated independently according to a probability distribution $\mathbf{P}(\mathbf{X},\mathbf{Y})$ on $\inputspace \times \outputspace$. Here, $\outputspace$ is the set of possible label combinations, i.e., the power set of $\setoflabels$. To ease notation, we define $\vec{y}_{i}$ as a binary vector $\vec{y}_{i}= (y_{i,1}, y_{i,2}, \ldots, y_{i,m} )$, in which $y_{i,j}=1$ indicates the presence (relevance) and $y_{i,j}=0$ the absence (irrelevance) of $\lambda_j$ in the labeling of $\vec{x}_{i}$. Under this convention, the output space is given by $\outputspace=\{0,1\}^{m}$. 
The goal in MLC  is to induce from $S$ a hypothesis $\vec{h}: \inputspace \longrightarrow \outputspace$ that correctly predicts the subset of relevant labels for unlabeled query instances $\vec{x}$. 

%:BR
The most straightforward and arguably simplest approach to tackle the MLC problem is \emph{binary relevance} (BR) learning. The BR method reduces a given multi-label problem with $m$ labels to $m$ \emph{binary classification} problems. More precisely, $m$ hypotheses $h_1, h_2, \ldots , h_m$ are induced, each of them being responsible for predicting the relevance of one label, using $\inputspace$ as an input space:
\begin{equation}
\label{eq:classifiers}
h_{j}: \inputspace \longrightarrow \{0,1\} 
\end{equation}
In this way, the labels are predicted independently of each other and no label dependencies are taken into account.

%:BR minimizes Hamming but not subset 0/1
In spite of its simplicity and the strong assumption of label independence, it has been shown theoretically and empirically that BR performs quite strong in terms of decomposable loss functions \cite{EykeBayes}, including the well-known \emph{Hamming loss}:
\begin{equation}
\label{eq:hamming}
L_H(\vec{y},\vec{h}(\vec{x})) = \frac{1}{m} \sum_{i=1}^{m} [\![ y_{i} \neq h_{i}(\vec{x}) ]\!]
\end{equation}
The Hamming loss averages the standard 0/1 classification error over the $m$ labels and hence corresponds to the proportion of labels whose relevance is incorrectly predicted. Thus, if one of the labels is predicted incorrectly, this accounts for an error of $\frac{1}{m}$. Another extension of the standard 0/1 classification loss is the \emph{subset 0/1 loss}:
\begin{equation}
\label{eq:zero-one}
L_{ZO}(\vec{y},\vec{h}(\vec{x})) =  [\![ \vec{y} \neq \vec{h}(\vec{x}) ]\!]
\end{equation}
Obviously, this measure is more drastic and already treats a mistake on a single label as a complete failure. 
%Only if all relevant labels have been predicted correctly, this loss function accounts for zero error. In contrast to Hamming, it accounts for an error of one, if only one label is predicted incorrectly. 
The necessity to exploit label dependencies in order to minimize the generalization error in terms of the subset 0/1 loss has been shown in \cite{EykeBayes}.

\section{Classifier Chains}\label{sec:CC}

%:CC
While following a similar setup as BR, classifier chains (CC) seek to capture label dependencies. CC learns $m$ binary classifiers linked along a chain, where each classifier deals with the binary relevance problem associated with one label. In the training phase, the feature space of each classifier in the chain is extended with the actual label information of all previous labels in the chain. For instance, if the chain follows the order $\lambda_{1} \rightarrow \lambda_{2}\rightarrow  \ldots \rightarrow \lambda_{m}$, then the classifier $h_{j}$ responsible for predicting the relevance of $\lambda_j$ is of the form
\begin{equation}
\label{eq:CC}
h_{j}: \, \inputspace \times \{0,1\}^{j-1} \longrightarrow \{0,1\} \enspace .
\end{equation}
The training data for this classifier consists of instances $(\vec{x}_{i}, y_{i,1}, \ldots, y_{i,j-1})$ labeled with $y_{i,j}$, that is, original training instances $\vec{x}_{i}$ supplemented by the relevance of the labels $\lambda_1, \ldots , \lambda_{j-1}$ preceding $\lambda_j$ in the chain. 

At prediction time, when a new instance $\vec{x}$ needs to be labeled, a label subset $\vec{y}=(y_1, \ldots , y_m)$ is produced by successively querying each classifier $h_j$. Note, however, that the inputs of these classifiers are not well-defined,  since the supplementary attributes $y_{i,1}, \ldots, y_{i,j-1}$ are not available. These missing values are therefore replaced by their respective predictions: $y_1$ used by $h_2$ as an additional input is replaced by $\hat{y}_1 = h_1(\vec{x})$,   $y_2$ used by $h_3$ as an additional input is replaced by $\hat{y}_2 = h_2(\vec{x}, \hat{y}_1)$, and so forth. Thus, the prediction $\vec{y}$ is of the form
$$
\vec{y} = \big( \, h_1(\vec{x}), \, h_2(\vec{x}, h_1(\vec{x})), \,  h_3(\vec{x}, h_1(\vec{x}), h_2(\vec{x}, h_1(\vec{x})))  , \ldots 
\big)
$$
% predicted. To fill the gaps in the input vector of the $j$-th classifier $h_j$, it takes the predictions of the $j-1$ previous classifiers $\{h_i\}_1^{j-1}$. Therefore, the classifiers are applied following the chain order, using the predictions of previous models as inputs.
%:ECC
Realizing that the order of labels in the chain may influence the performance of the classifier, and that an optimal order is hard to anticipate,  the authors in \cite{ClassifierChainsML} propose the use of an ensemble of CC classifiers. This approach combines the predictions of different random orders and, moreover, uses a different sample of the training data to train each member of the ensemble. \emph{Ensembles of classifier chains} (ECC) have been shown to increase predictive performance over CC by effectively using a simple voting scheme to aggregate predicted relevance sets of the individual CCs: For each label $\lambda_j$, the proportion $\hat{w}_j$ of classifiers predicting $y_j = 1$ is calculated. Relevance of $\lambda_j$ is then predicted by using a threshold $t$, that is, $\hat{y}_j = [\![ \hat{w}_j \geq t  ]\!]$.

%Given a threshold $t$ the ensemble prediction is evaluated using a threshold function: 
%
%\begin{equation}
%\label{eq:thresholdFunction}
%f_t(\hat{\vec{w}}) = 
%\begin{cases} 
% 1 & \hat{w}_j \geq t \\
% 0 & \hat{w}_j < t \\
%\end{cases} \\[0.5ex]
%\end{equation}
%  

\section{Attribute Noise in Classifier Chains}\label{sec:IID}
 
%:violation
The learning process of CC violates a key assumption of supervised learning, namely the assumption that the training data is representative of the test data in the sense of being identically distributed. This assumption does not hold for the chained classifiers in CC: While using the \emph{true} label data $y_j$ as input attributes during the training phase, this information is replaced by \emph{estimations} $\hat{y}_j$ at prediction time. Needless to say, $y_j$ and $\hat{y}_j$ are not guaranteed to follow the same distribution; on the contrary, unless the classifiers produce perfect predictions, these distributions are likely to differ in practice (in particular, note that the $\hat{y}_j$ are deterministic predictions whereas the $y_j$ normally follow a non-degenerate probability distribution).
 
From the point of view of the classifier $h_j$, which uses the labels $y_1, \ldots , y_{j-1}$ as additional attributes, this problem can be seen as a problem of \emph{attribute noise}. More specifically, we are facing the ``clean training data vs.\ noisy test data'' case, which is one of four possible noise scenarios that have been studied quite extensively in \cite{ZhuNoise}. For CC, this problem appears to be vital: Could it be that the additional label information, which is exactly what CC seeks to exploit in order to gain in performance (compared to BR), eventually turns out to be a source of impairment? Or, stated differently, could the additional label information perhaps be harmful rather than useful?

%:introduce factors
This question is difficult to answer in general. In particular, there are several factors involved, notably the following: 
%The above discussion suggest, that there are several effects interfering with each other. Additionally, revising the experimental results of several authors and our own refering to CC draws a mixed picture. It seems, that there are situations in which CC is able to gain performance over BR and others in which this is not the case. In order to settle the situation, the following determining factors have to be distinguished, which all affect the performance of CC:
%
\begin{itemize}
\item \emph{The length of the chain}: The larger the number $j-1$ of preceding classifiers in the chain, the higher is the potential level of attribute noise for a classifier $h_j$. For example, if prediction errors occur independently of each other with probability $\epsilon$, then the probability of a noise-free input is only $(1-\epsilon)^{j-1}$. More realistically, one may assume that the probability of a mistake is not constant but will increase with the level of attribute noise in the input. Then, due to the recursive structure of CC, the probability of a mistake will be multiplied and increase even more rapidly along the chain. 

\item \emph{The order of the chain}: Since some labels might be inherently more difficult to predict than others, the order of the chain will play a role, too. In particular, it would be advantageous to put simpler labels in the beginning and harder ones more toward the end of the chain.

\item \emph{The accuracy of the binary classifiers}: The level of attribute noise is in direct correspondence with the accuracy of the binary classifiers along the chain. More specifically, these classifiers determine the input distributions in the test phase. If they are perfect, then the training distribution equals the test distribution, and there is no problem.  Otherwise, however, the distributions will differ.

\item \emph{The dependency among labels}: Perhaps most interestingly, a (strong enough) dependence between labels is a prerequisite for both, an improvement and a deterioration through chaining. In fact, CC cannot gain (compared to BR) in case of no label dependency. In that case, however, it is also unlikely to lose, because a classifier $h_j$ will most likely\footnote{The possibility to ignore parts of the input information does of course also depend on the type of base classifier used.} ignore the attributes $y_1, \ldots , y_{j-1}$. Otherwise, in case of pronounced label dependence, it will rely on these attributes, and whether or not this is advantageous will depend on the other factors above.  

%Most interestingly, the label dependence, which CC tries to exploit is also inherently the reason, why a sub model should at all suffer from the attribute noise introduced by its predecessor model. If there would be no label dependence between consequent labels $\lambda_{j-1}$ and $\lambda_{j}$ in the chain, it is likely, that $h_{j}$ will not take $h_{j-1}$'s prediction into account. This, of cause, also depends on the model class. \emph{Decision trees} \cite{c45} e.g., will probably not even use this label information at all, since they conduct an integrated feature selection. But also linear models like \emph{logistic regression} \cite{logit} will most probably assign a very low coefficient to this feature. In contrast, if there is a high dependency between these to labels, or even a full correlation, the induced sub model will most certainly fully rely on the previous label information and will fail, whenever the previous model fails.
\end{itemize}
%
%The level of testing noise directly depends on the accuracy of the previous classifiers in the chain. What we can expect is that in those domains in which the accuracy is low, CC could have a poor performance. It is important to emphasize that testing noise can be much worse than attribute noise. Almost every inductive learning algorithm has a mechanism for handling attribute noise, whenever the  noise level is not to high. However, CC's binary models cannot deal in any way with the testing noise described before. Simply, because it is not reflected by the training data. 
%
%Another factor that can have an impact in the performance of CC is the length of the chains. As it was explained at the beginning of this section, the chains have as many classifiers as labels. When the number of labels is large, the probability that one of the binary classifiers along the chain make a mistake is bigger. Every misclassification implies a higher level of testing noise for the following classifiers. Thus, the level of testing noise is likely to increase along the chain.
%
In the following, we present two experimental studies that are meant to illustrate the above issues. Based on our discussion so far and these experiments, two modifications of CC will then be introduced in the next sections, both of them with the aim to alleviate the problems outlined above. 

\begin{figure}[t!]
\begin{center}
\includegraphics[scale=0.52]{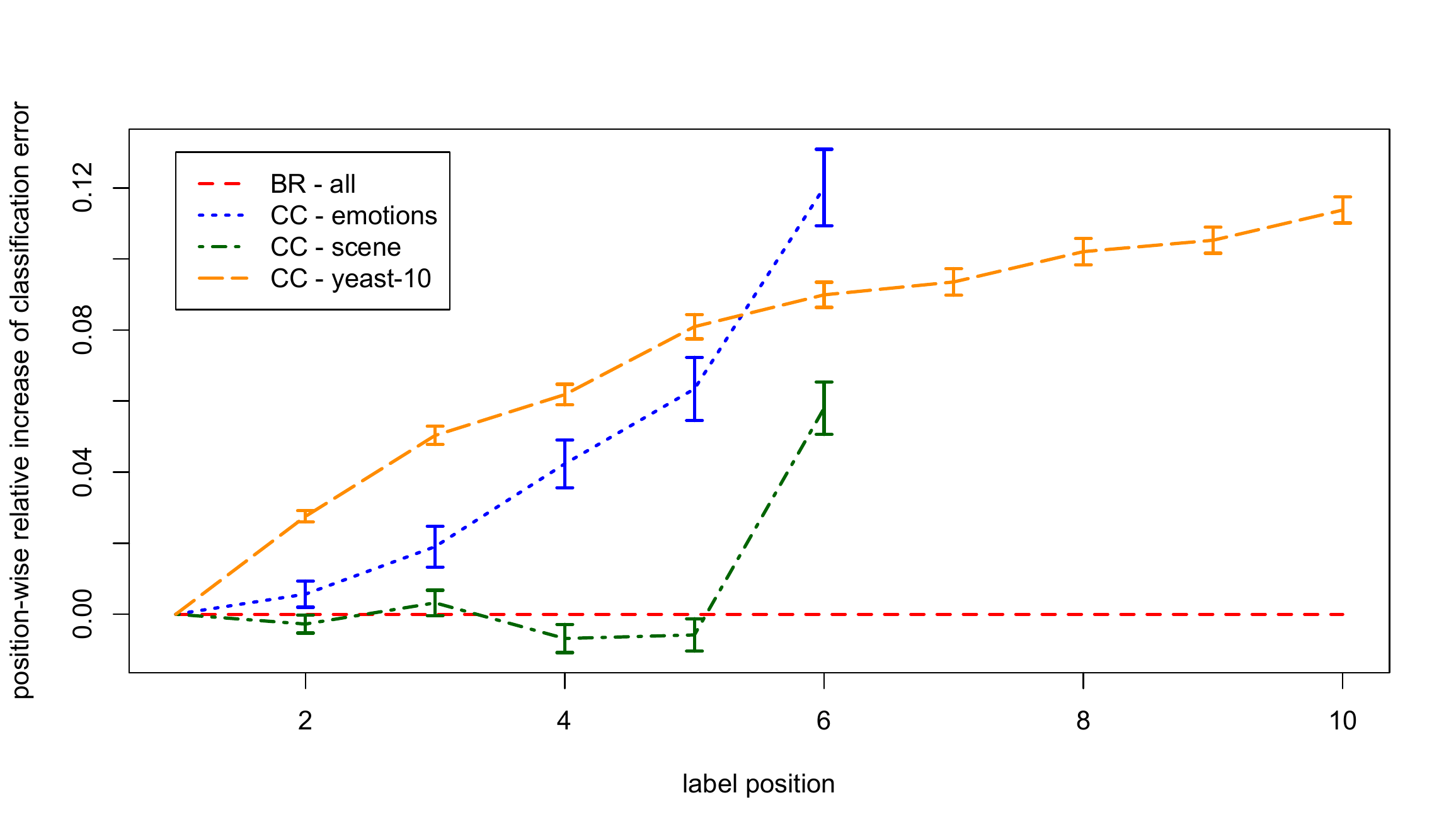}
\caption{Results of the first experiment: position-wise relative increase of classification error (mean plus standard error bars). The \emph{yeast-10} data set used here is a reduced yeast data set containing only the ten most frequent labels and their instances.}
\label{lateError}
\end{center}
\end{figure}

\subsection{First Experiment}

Our intuition is that attribute noise in the test phase can produce a propagation of errors through the chain, thereby affecting the performance of the classifiers depending on their position in the chain. More specifically, we expect classifiers in the beginning of the chain to systematically perform better than classifiers toward the end. In order to verify this conjecture, we perform the following simple experiment: We train a CC classifier on 500 randomly generated label orders. Then, for each label order and each position, we compute the performance of the classifier on that position in terms of the relative increase of classification error compared to BR. Finally, these errors are averaged \emph{position-wise} (not label-wise).  For this experiment, we used three standard MLC benchmark data sets whose properties are summarized in Table \ref{tab:data sets} (shown in Section \ref{sec:NS}).
%The classifiers at first position are equal to those of BR learning. Their predictions are used by the classifiers at the second position, and so on.

The results in Figure \ref{lateError} clearly confirm our expectations. In two cases, CC starts to lose immediately, and the loss increases with the position. In the third case, CC is able to gain on the first positions but starts to lose again later on. 
%shows the accuracy gain as a function of the position. The results support our claims, that there are cases, where CC suffers from the violation of the i.i.d. assumption, but there is also one case (scene), where CC can gain performance, at least for the first five labels. 

%
\begin{figure}[t!]
\begin{center}
\includegraphics[scale=0.45]{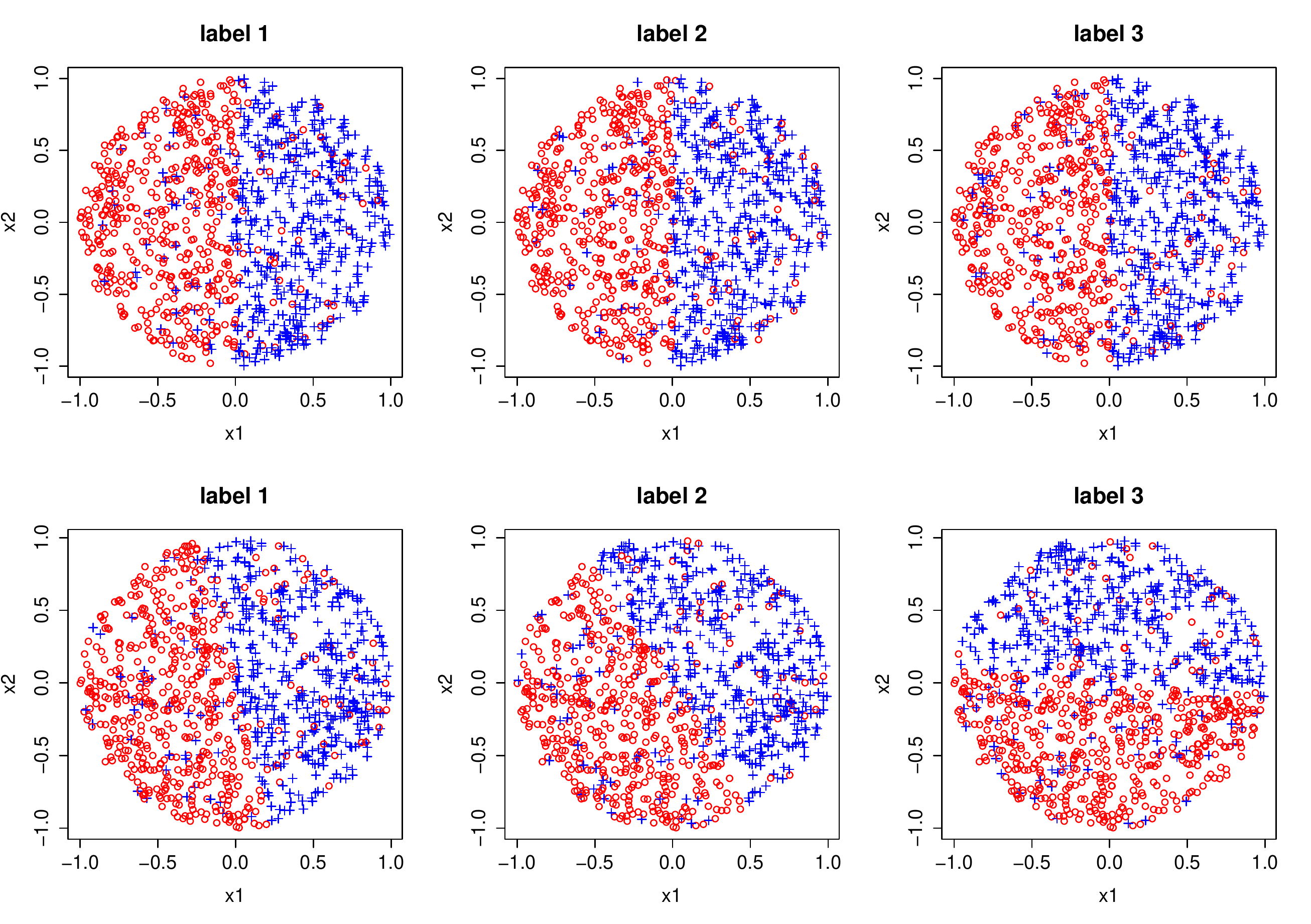}
\caption{Example of synthetic data: the top three labels are generated using $\tau=0$, the three at the bottom with $\tau=1$.}
\label{synth1_example}
\end{center}
\end{figure}

\subsection{Second Experiment}
\label{sec:syn2}

%:synthetic data sets
In a second experiment, we use a synthetic setup that was proposed in \cite{LabelDependenceML} to analyze the influence of label dependence. The input space $\inputspace$ is two-dimensional and the underlying decision boundary for each label is linear in these inputs. More precisely, the model for each label is defined as follows:
\begin{equation}
\label{eq:synth1}
h_{j}(\vec{x}) = 
\begin{cases} 
  1 & a_{j,1}x_1+a_{j,2}x_2 \ge 0\\
  0 & \text{otherwise} \\
\end{cases} \\
\end{equation}
%
%where $j \in \{1, ...,m\}$. 
The input values are drawn randomly from the unit circle. The parameters $a_{j,1}$ and $a_{j,2}$ for the $j$-th label are set to  
\begin{equation}
\label{eq:synth1Param}
a_{j,1} = 1-\tau r_1, \enspace a_{j,2} = \tau r_2 \enspace ,
\end{equation}
with $r_1$ and $r_2$ randomly chosen from the unit interval. Additionally, random noise is introduced for each label by independently reversing a label with probability $\pi=0.1$. 
\begin{figure}[t!]
\begin{center}
\includegraphics[scale=0.45]{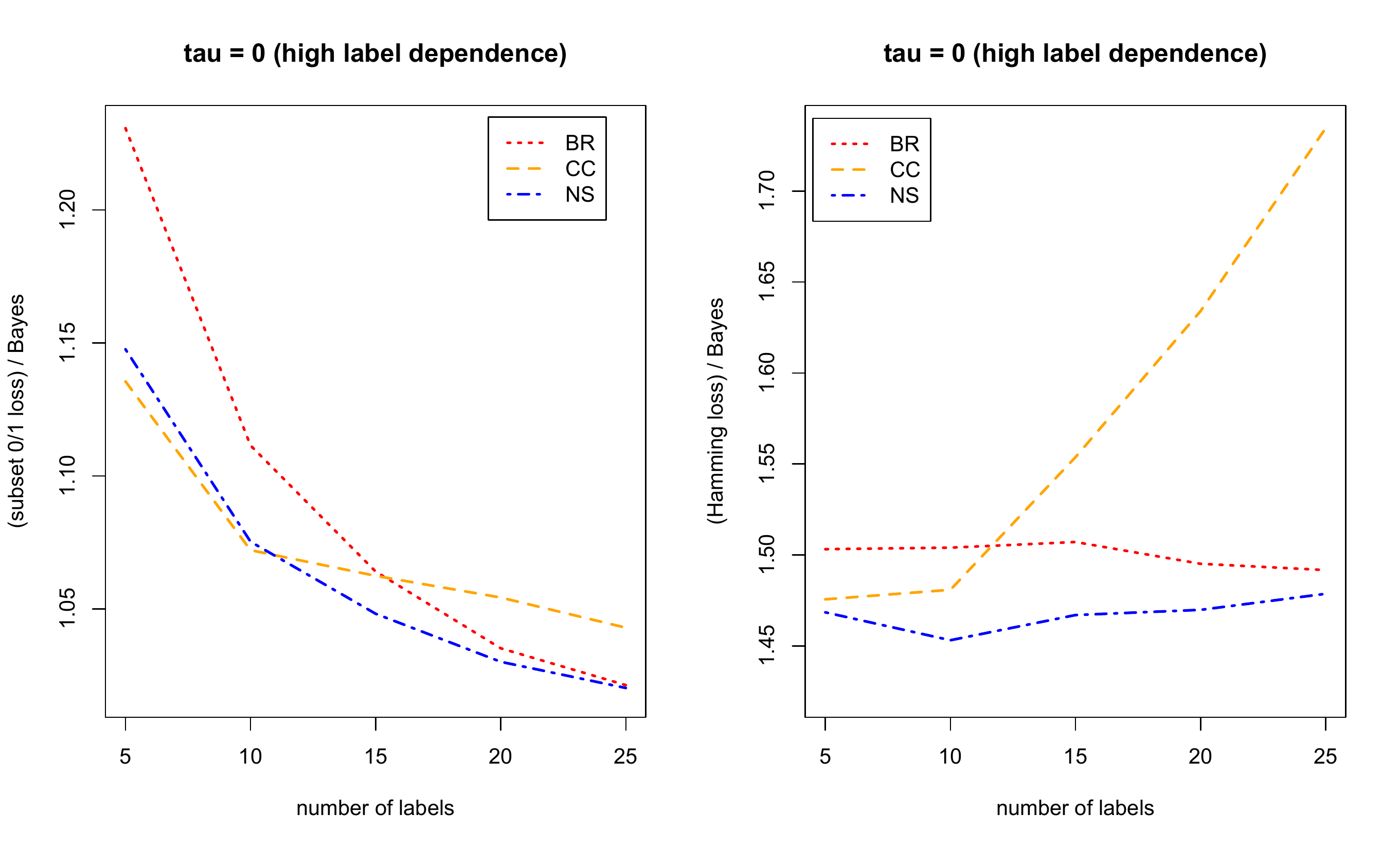}
\includegraphics[scale=0.45]{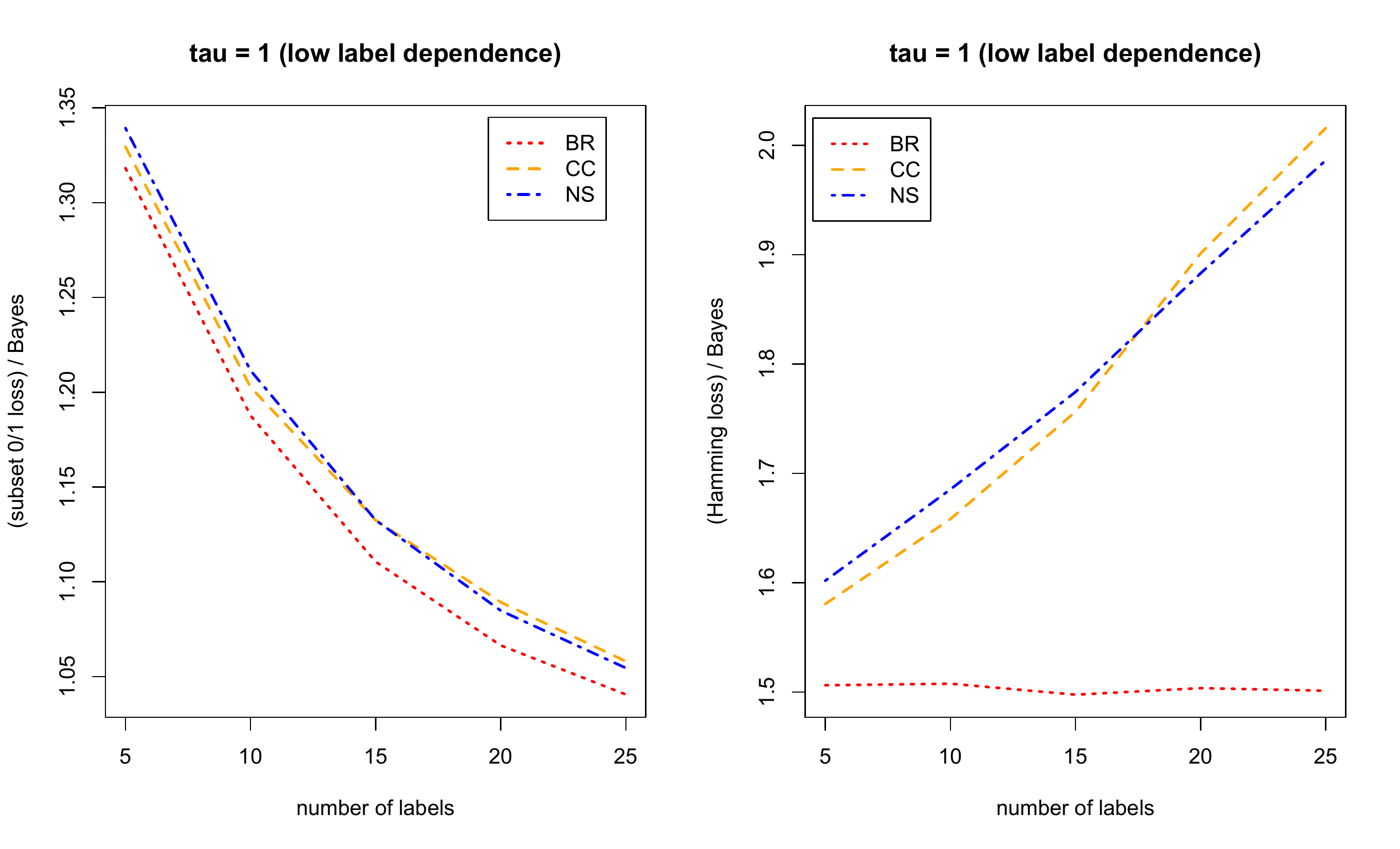}
\caption{Results of the second experiment for $\tau=0$ (top---high label dependence) and $\tau=1$ (bottom---low label dependence).}
\label{resultsSynth1}
\end{center}
\end{figure}
Obviously, the level of label dependence can be controlled by the parameter $\tau$. Figure \ref{synth1_example} shows two example data sets with three labels. The first  one (pictures on the top) is generated with $\tau=0$, the second one (bottom) with $\tau=1$. As can be seen, the label dependence is quite strong in the first case, where the model parameters (\ref{eq:synth1Param}) are the same for each label. For the second case, the model parameters are different for each label. There is still label dependence, but certainly less pronounced.

%:experimental setup
For different label cardinalities $m \in \{5,10,15,20,25\}$, we run 10 repetitions of the following experiment: We created 10 different random model parameter sets (two for each label) and generated 10 different training sets, each consisting of 50 instances. For each training set, a model is learnt and evaluated (in terms of Hamming and subset 0/1 loss) on an additional data set comprising 1000 instances. 

%:bayes optimal error
%Since we know the true underlaying models, we also know the Bayes optimal error in terms of each loss. For Hamming loss, this is equal to $\pi=0.1$, which is the level of class noise we introduced for each label. For subset 0/1 loss however, it additionally depends on the number of labels in the data set. The Bayes optimal error in this case is $1-(1-\pi)^m$ which is strictly increasing with the number of labels. 
%
%:explaining the plots
%Standard error bars are omitted, because the standard errors in this experiment have been very small in all cases. To be able to visually distinguish the different methods within the plots, it was necessary to proceed in the same way as in the previous experiment. All results have been compared to the results of BR. The plots only show the relative difference to BR of each method.

%:interpreting the results
Figure \ref{resultsSynth1} summarizes the results in terms of the average loss divided by the corresponding Bayes loss (which can be computed since the data generating process is known); thus, the optimum value is always 1. Apart from BR and CC, we already include the performance curve for the method to be introduced in the next section (NS); this should be ignored for now. Comparing BR and CC, the big picture is quite similar to the previous experiment: The performance of CC tends to decrease relative to BR with an increasing number of labels. In the case of low label dependence, this can already be seen for only five labels. The case of high label dependence is more interesting: While CC seems to gain from exploiting the dependency for a small to moderate number of labels, it cannot extend this gain to more than 15 labels.

%On Label Dependence and Loss Minimization in Multi-Label Classification. Page 24: Interestingly, CC is not better than BR in terms of Hamming loss in the case of the same structural parts. Moreover, the standard errors of the Hamming loss are for CC indifferent to the number of labels. For t = 1, its performance decreases if the number of labels increases. However, it performs much better with respect to subset 0/1 loss, and its behavior is similar to BR in this case. These results can be interpreted as follows. For the same structural parts, CC tends to build a model based on values of previous labels. In the prediction phase, however, once the error is made, it will be propagated along a chain. From this point of view, its behavior is similar to using for all labels a base classifier that has been learned on the first label. That is why standard errors do not change in the case of Hamming loss. This behavior gives a small advantage for subset 0/1 loss, as the predictions become more homogeneous."

\section{Nested Stacking}\label{sec:NS}

A first very simple idea to mitigate the problem of attribute noise in CC is to let a classifier $h_j$ use predicted labels $\hat{y}_1, \ldots , \hat{y}_{j-1}$ as supplementary attributes for training instead of the true labels  $y_1, \ldots , y_{j-1}$. This way, one could make sure that the data distribution is the same for training and testing. Or, stated differently, the situation faced by a classifier during training does indeed equal the one it will encounter later on at prediction time. 
Since then a classifier is trained on the predictions of other classifiers, this approach fits the stacked generalization learning paradigm \cite{WolpertStacked}, also simply known as \emph{stacking}.

\subsection{Stacking versus Nested Stacking}
%STACKING
The idea of stacking has already been used in the context of MLC by Godbole and Sharawagi \cite{GodboleDiscriminative}.
% to overcome the label independence problem of BR. They apply the stacked generalization learning paradigm \cite{WolpertStacked}, also known simply as stacking, in the context of multi-label classification. 
In the learning phase, their method builds a stack of two groups of classifiers. The first one is formed by the standard BR classifiers: 
$$
\vec{h}^{1}(\vec{x})=(h_{1}^{1}(\vec{x}), \ldots, h_{m}^{1}(\vec{x}) ) \, . 
$$
On a second level, also called meta-level, another group of binary models (again one for each label) is learnt, but these classifiers consider an augmented feature space that includes the binary outputs of all models of the first level: 
$$
\vec{h}^{2}(\vec{x},\vec{y}')=(h_{1}^{2}(\vec{x},\vec{y}'),  \ldots, h_{m}^{2}(\vec{x},\vec{y}') ) \, , 
$$
where $\vec{y}'=\vec{h}^{1}(\vec{x})$.
The idea is to capture label dependencies by learning their relationships in the meta-level step. In the test phase, the final predictions are the outputs of the meta-level classifiers, $\vec{h}^{2}(\vec{x})$, using the outputs of $\vec{h}^{1}(\vec{x})$ exclusively to obtain the values of the augmented feature space. 

Mimicking the chain structure of CC, our variant of stacking is a \emph{nested} one: Instead of a two-level architecture as in standard stacking, we obtain a nested hierarchy of stacked (meta-)classifiers. Hence, we call it \emph{nested stacking} (NS). Moreover, each of these classifiers is only trained on a subset of the predictions of other classifiers. Like in CC, $m$ models need to be trained in total, while $2m$ models are trained in standard stacking. 

\subsection{Out-of-Sample versus Within-Sample Training}

To make sure that the distribution of the labels 
$\hat{y}_1, \ldots , \hat{y}_{j-1}$, which are used as supplementary attributes by the classifier $h_j$, is indeed the same at training and prediction time, these labels should be produced by means of an out-of-sample prediction procedure. For example, an internal leave-one-out cross validation procedure could be implemented for this purpose. 
%, because this way, the training data used to predict the label of a given instance at training time is most similar to the data available to create the final classifier used in at testing time. In fact, they only differ by one instance. 

Needless to say, a procedure of that kind is computationally complex, even for classifiers that can 
%We are aware of the fact, that an internal out-of-sample procedure like this in general comes with high computational cost. Nevertheless, there are classifiers, which are able to 
be trained and ``detrained'' incrementally (such as incremental and decremental support vector machines \cite{cauwenberghs2001incremental}). In our current version of NS, we therefore implement a simple within-sample strategy. In several experimental studies, we found this strategy to perform almost as good as out-of-sample training, while being significantly faster. In fact, methods such as logistic regression, which are not overly flexible, are hardly influenced by excluding or including a single example.

\subsection{A First Experiment}

To get a first impression of the performance of NS, we return to the experiment in Section \ref{sec:syn2}. As can be seen in Figure \ref{resultsSynth1}, NS does indeed gain in comparison to CC with an increasing number of labels; only if the labels are few, CC is still a bit better. This tendency is more pronounced in the case of strong label dependency, whereas the differences are rather small if label dependence is low.

To explain the competitive performance of CC if the number of labels is small, note that replacing ``clean'' training data $y_1, \ldots , y_{j-1}$ by possibly more noisy data $\hat{y}_1, \ldots , \hat{y}_{j-1}$, as done by NS, may not only have the positive effect of making the training data more authentic. In fact, it may also make the problem of learning $h_j$ more difficult (because the dependency $y_1, \ldots , y_{j-1} \rightarrow y_j$ might be ``easier'' than the dependency $\hat{y}_1, \ldots , \hat{y}_{j-1} \rightarrow y_j$).
Apparently, this effect plays an important role if the number of labels is small, whereas the positive effect dominates for longer label chains. 

%may be explained by a discussion on the following two apparent effects. NS equalizes the distributions of training and test data as described above, which results in models, which are trained to deal with noisy predictions of the previous labels. This positive effect predominates in the case of many labels, when propagated errors have a strong impact (see Section~\ref{sec:IID}). However, introducing noise in the training data can also have a negative effect for NS, because it blurres the original relation between the input variables and the labels. This yields a more difficult learning problem for each label. The results suggest, that the later effect plays an important role, if the number of labels is small.

\subsection{Subset Correction}
\label{sec:subset_correction}

%Before comparing CC and NS on benchmark data sets, we want to suggest another procedure which is rather general to all MLC learners and which is not restricted to CC and NS. In the realm of subset 0/1 loss and in contrast to other loss functions like Hamming, the prediction of an multi-label classifier (a subset of relevant labels) is treated as a unity. Instead, Hamming and most other losses, treat this subset as a collection of classical binary predictions, which are, in one way or the other aggregated into an overall evaluation. Stated differently, if subset 0/1 loss is used, one can see MLC as a special case of standard multi-class classification, where there are $2^m$ classes, each of which is representing one of the possible subsets of labels. This point of view motivated a procedure, which we call \emph{subset correction}. The idea is quite simple. In standard multi-class classification training data should at least contain one example for every class. If not, it is questionable, why a learner should be able to predict a class, which it has never seen before. Mapped to the MLC case, this suggests, that a multi-label classifier is only allowed to predict "classes" (label subsets), which have been observed at least once in the training data. This also is another way to indirectly take into account label dependencies. Unknown label subsets are considered to be invalid.

Our second modification is motivated by the observation that the number of label combinations that are commonly observed in MLC data sets is only a tiny fraction of the total number $|\outputspace| = 2^{m}$ of possible subsets; see Table~\ref{tab:data sets}, which reports the value $|\outputspace_D|2^{-m}$, where $\outputspace_D$ is the set of unique label combinations contained in the data $D$, as the ``observation rate'' in the last column. Moreover, if a label combination $\vec{y}$ has an occurrence probability of $\epsilon>0$, then the probability that it has never be seen in a data set of size $n$ reduces to $(1-\epsilon)^n$. Thus, by contraposition, one may argue that such a label combination is indeed unlikely to exist at all (at least for large enough $n$). 

% With a maximum total number of possible subset of $2^{159}$ (bibtex), the number of observed label subsets in the training data is only a very small fraction.

Our idea of ``subset correction'', therefore, is to restrict a learner to the prediction of label combinations whose existence is testified by the (training) data. More precisely, let $\outputspace_S$ denote the set of label subsets $\vec{y}$ that have be seen in the training data $S$. Then, given a prediction $\vec{\hat{y}}$ produced by a classifier $\vec{h}$, this prediction is replaced by the ``most similar'' subset $\vec{y}^* \in \outputspace_S$:
\begin{equation}\label{eq:ssc}
\vec{y}^* \in \argmin\limits_{\vec{y}' \in \outputspace_S} L_H(\vec{\hat{y}},\vec{y}') 
\end{equation}
Thus, $\vec{y}^*$ is eventually returned as a prediction instead of $\vec{\hat{y}}$. 
If the minimum in (\ref{eq:ssc}) is not unique, those label combinations with higher frequency in the training data are preferred.

In principle, the Hamming loss could of course be replaced by other MLC loss functions in (\ref{eq:ssc}). Its use here is mainly motivated by the fact, that it is used for a similar purpose, namely decoding, in the framework of \emph{error correcting output codes} (ECOC). As such, it has been applied in multi-class classification \cite{Dietterich_95} and lately also in MLC \cite{Kajdanowicz_12},\cite{Fuernkranz_12}.

\section{Nested Stacking versus Classifier Chains}\label{sec:NSexperiments}

In this section, we compare NS and CC, both with and without subset correction, on real MLC benchmark data. As can be seen in Table~\ref{tab:data sets}, the data sets differ quite significantly in terms of the number of attributes, examples, labels, cardinality (number of labels per example) and the observation rate. 

\begin{table}[t]
\caption{Properties of the data sets used in the experiments.}
\begin{center}
\begin{tabular}{lrrrrrr}
\hline
Data set & Attributes & Examples & Labels & Cardinality & Observation Rate\\
\hline
bibtex    & 1836  & 7395  & 159   & 2.40 & 3.9E-45\\
emotions    & 72    & 593   & 6     & 1.87 & 4.0E-1\\
enron       & 1001  & 1702  & 53    & 3.38 & 8.3E-14\\
genbase     & 1185  & 662   & 27    & 1.25 & 2.3E-7\\
image       & 135   & 2000  & 5     & 1.24 & 6.0E-1\\
mediamill   & 120   & 5000  & 101   & 4.27 & 2.5E-27\\
medical     & 1449  & 978   & 45    & 1.25 & 2.6E-12\\
reuters     & 243   & 7119  & 7     & 1.24 & 1.9E-1\\
scene       & 294   & 2407  & 6     & 1.07 & 2.3E-1\\
slashdot    & 1079  & 3782  & 22    & 1.18 & 3.7E-5\\
yeast       & 103   & 2417  & 14    & 4.24 & 1.2E-2\\
\hline
\end{tabular}
\end{center}
\label{tab:data sets}
\end{table}

%:base learner
Logistic regression was used as a base learner for binary prediction in all MLC methods \cite{liblr08}. 
%The regularization parameter $C$  was established for each \emph{individual} binary classifier by performing a grid search over the values $C \in \{ 10^{-3}, 10^{-2}, \ldots, 10^3 \}$, 
%\{10^{p} \mid p \in [-3, \ldots, 3]\}$, 
%optimizing the accuracy estimated by means of a 2-fold cross validation repeated 5 times. This guarantees the binary classifiers for a particular label of both methods to be exactly the same when their respective feature spaces coincide\footnote{This only happens for the first classifier of the chain in this comparison.}. 
%Moreover, the label orders of both approaches are the same in every experiment. 
Unlike \cite{ClassifierChainsML}, we do not apply any threshold selection procedure; instead, we simply used $t=0.5$ for deciding the relevance of a label. In fact, our goal is to study the behavior of CC and NS without the influence of other factors that may bias the results. 
%Therefore, the differences between both approaches are solely due to the different feature spaces used. 

Since CC's main goal is to detect conditional label dependence, we used example-based metrics for evaluation. In addition to Hamming and subset 0/1 loss introduced earlier, we also applied the $F_1$ and Jaccard index defined, respectively, as follows (note that these are accuracy measures instead of loss functions):
\begin{equation}
\label{eq:f1}
F_{1}(\vec{y},\vec{h}(\vec{x})) = \frac{ 2 \sum_{i=1}^{m} [\![ y_{i} = 1 \textrm{ and } h_{i}(\vec{x}) = 1 ]\!] }{ \sum_{i=1}^{m} ( [\![ y_{i}=1 ]\!] + [\![ h_{i}(\vec{x}) = 1 ]\!] ) }
\end{equation}

\begin{equation}
\label{eq:jaccard}
Jaccard(\vec{y},\vec{h}(\vec{x})) = \frac{\sum_{i=1}^{m} [\![ y_{i} = 1 \textrm{ and } h_{i}(\vec{x}) = 1 ]\!] }{ \sum_{i=1}^{m} [\![ y_{i}=1 \textrm{ or } h_{i}(\vec{x}) = 1 ]\!] }
\end{equation}
The value for a test set is defined as the average over all instances. The scores reported in Tables~\ref{tab:NSvsCC} and~\ref{tab:NSscvsCCsc} were estimated by means of 10-fold cross-validation, repeated three times. We used a paired t-test for establishing statistical significance on each data set.

\begin{table}[t]
\caption{Experimental results of NS and CC on benchmark data sets. $\upuparrows$  ($\downdownarrows$) means that NS is significantly better (worse) than CC at level $p<0.01$ ($\uparrow$ and $\downarrow$ at level $p<0.05$) in a paired t-test.}
\centering
{\scriptsize
\begin{tabular}{lc@{\hspace{1.5em}}c@{\hspace{1em}}cl@{\hspace{2em}}c@{\hspace{1em}}cl} 
%\hline
%\vspace{-1.22pt}
 &  & \multicolumn{2}{c}{$F_{1}$} & & \multicolumn{2}{c}{\textsc{Jaccard Index}} & \\
 \hline
  & $m$  & CC & NS & & CC & NS & \\
 \hline 
bibtex& 159   & 0.1697{\scriptsize$\pm$.0071} & 0.1747{\scriptsize$\pm$.0077} &$\upuparrows$    & 0.1098{\scriptsize$\pm$.0060} & 0.1133{\scriptsize$\pm$.0064}& $\upuparrows$    \\
emotions &6   & 0.5883{\scriptsize$\pm$.0534} & 0.6028{\scriptsize$\pm$.0500} &$\uparrow$     & 0.5003{\scriptsize$\pm$.0521} & 0.5144{\scriptsize$\pm$.0514}& $\uparrow$     \\
enron& 53     & 0.3483{\scriptsize$\pm$.0191} & 0.3729{\scriptsize$\pm$.0214} &$\upuparrows$    & 0.2474{\scriptsize$\pm$.0163} & 0.2693{\scriptsize$\pm$.0178}& $\upuparrows$    \\
genbase &27   & 0.9863{\scriptsize$\pm$.0090} & 0.9854{\scriptsize$\pm$.0085} &$\downarrow$   & 0.9804{\scriptsize$\pm$.0115} & 0.9789{\scriptsize$\pm$.0109}& $\downarrow$   \\
image   &5    & 0.5556{\scriptsize$\pm$.0284} & 0.4780{\scriptsize$\pm$.0299} &$\downdownarrows$  & 0.5196{\scriptsize$\pm$.0271} & 0.4460{\scriptsize$\pm$.0278}& $\downdownarrows$  \\
mediamill&101   & 0.5326{\scriptsize$\pm$.0054} & 0.5619{\scriptsize$\pm$.0053} &$\upuparrows$    & 0.4280{\scriptsize$\pm$.0052} & 0.4459{\scriptsize$\pm$.0052}& $\upuparrows$    \\
medical & 45  & 0.6462{\scriptsize$\pm$.0331} & 0.6444{\scriptsize$\pm$.0340} &         & 0.5828{\scriptsize$\pm$.0343} & 0.5804{\scriptsize$\pm$.0356}&          \\
reuters    &7   & 0.8599{\scriptsize$\pm$.0128} & 0.8570{\scriptsize$\pm$.0116} &$\downdownarrows$  & 0.8336{\scriptsize$\pm$.0138} & 0.8302{\scriptsize$\pm$.0129}& $\downdownarrows$  \\
scene&  6     & 0.5969{\scriptsize$\pm$.0403} & 0.6031{\scriptsize$\pm$.0348} &         & 0.5745{\scriptsize$\pm$.0405} & 0.5766{\scriptsize$\pm$.0344}&          \\
slashdot  &22   & 0.3278{\scriptsize$\pm$.0185} & 0.3259{\scriptsize$\pm$.0186} &         & 0.2747{\scriptsize$\pm$.0176} & 0.2726{\scriptsize$\pm$.0180}&          \\
yeast&  14    & 0.5836{\scriptsize$\pm$.0182} & 0.6068{\scriptsize$\pm$.0172} &$\upuparrows$    & 0.4848{\scriptsize$\pm$.0198} & 0.4990{\scriptsize$\pm$.0183}& $\upuparrows$    \\
\hline
\vspace{0.1cm}\\
 &  & \multicolumn{2}{c}{\textsc{Hamming Loss}} & & \multicolumn{2}{c}{\textsc{Subset 0/1 Loss}} & \\
 \hline
  & $m$ & CC & NS & & CC & NS \\
 \hline 
bibtex& 159   & 0.0724{\scriptsize$\pm$.0020} & 0.0672{\scriptsize$\pm$.0016}&$\upuparrows$ & 0.9837{\scriptsize$\pm$.0052} & 0.9833{\scriptsize$\pm$.0052}&          \\ 
emotions &6   & 0.2367{\scriptsize$\pm$.0268} & 0.2169{\scriptsize$\pm$.0253}&$\upuparrows$ & 0.7578{\scriptsize$\pm$.0575} & 0.7477{\scriptsize$\pm$.0633}&          \\
enron & 53    & 0.1233{\scriptsize$\pm$.0051} & 0.1050{\scriptsize$\pm$.0051}&$\upuparrows$ & 0.9565{\scriptsize$\pm$.0135} & 0.9510{\scriptsize$\pm$.0133}&$\uparrow$      \\
genbase &27   & 0.0019{\scriptsize$\pm$.0011} & 0.0020{\scriptsize$\pm$.0010}&$\downarrow$  & 0.0408{\scriptsize$\pm$.0211} & 0.0443{\scriptsize$\pm$.0213}&$\downarrow$    \\
image & 5     & 0.2104{\scriptsize$\pm$.0127} & 0.1962{\scriptsize$\pm$.0119}&$\upuparrows$ & 0.5857{\scriptsize$\pm$.0269} & 0.6468{\scriptsize$\pm$.0249}&$\downdownarrows$ \\
mediamill&101 & 0.0303{\scriptsize$\pm$.0004} & 0.0291{\scriptsize$\pm$.0004}&$\upuparrows$ & 0.8752{\scriptsize$\pm$.0049} & 0.8969{\scriptsize$\pm$.0048}&$\downdownarrows$ \\
medical & 45  & 0.0248{\scriptsize$\pm$.0031} & 0.0249{\scriptsize$\pm$.0031}&        & 0.5890{\scriptsize$\pm$.0425} & 0.5934{\scriptsize$\pm$.0463}&          \\
reuters & 7   & 0.0506{\scriptsize$\pm$.0046} & 0.0483{\scriptsize$\pm$.0043}&$\upuparrows$ & 0.2454{\scriptsize$\pm$.0173} & 0.2499{\scriptsize$\pm$.0175}&$\downarrow$    \\
scene & 6     & 0.1470{\scriptsize$\pm$.0143} & 0.1397{\scriptsize$\pm$.0124}&$\upuparrows$ & 0.4918{\scriptsize$\pm$.0434} & 0.5019{\scriptsize$\pm$.0355}&$\downarrow$    \\
slashdot& 22  & 0.0908{\scriptsize$\pm$.0027} & 0.0913{\scriptsize$\pm$.0028}&$\downarrow$  & 0.8652{\scriptsize$\pm$.0185} & 0.8678{\scriptsize$\pm$.0198}&          \\
yeast & 14    & 0.2242{\scriptsize$\pm$.0093} & 0.2069{\scriptsize$\pm$.0087}&$\upuparrows$ & 0.8104{\scriptsize$\pm$.0229} & 0.8469{\scriptsize$\pm$.0231}&$\downdownarrows$ \\
\hline
\end{tabular}
}
\label{tab:NSvsCC}
\end{table}

\begin{table}[t]
\caption{Experimental results of $\text{NS}_{SC}$ and $\text{CC}_{SC}$ on benchmark data sets. $\upuparrows$ ($\downdownarrows$) means that $\text{NS}_{SC}$ is significantly better (worse) than $\text{CC}_{SC}$ at level $p<0.01$ ($\uparrow$ and $\downarrow$ at level $p<0.05$) in a paired t-test.}
\centering
{\scriptsize
\begin{tabular}{lc@{\hspace{1.5em}}c@{\hspace{1em}}cl@{\hspace{2em}}c@{\hspace{1em}}cl} 
%\hline
%\vspace{-1.22pt}
 &  & \multicolumn{2}{c}{$F_{1}$} & & \multicolumn{2}{c}{\textsc{Jaccard Index}} & \\
 \hline
  & $m$  & $\text{CC}_{SC}$ & $\text{NS}_{SC}$ & & $\text{CC}_{SC}$ & $\text{NS}_{SC}$ & \\
 \hline 
bibtex    & 159   & 0.2026{\scriptsize$\pm$.0119} & 0.2090{\scriptsize$\pm$.0113} & $\upuparrows$   & 0.1528{\scriptsize$\pm$.0099} & 0.1582{\scriptsize$\pm$.0100}& $\upuparrows$    \\
emotions  & 6     & 0.5905{\scriptsize$\pm$.5905} & 0.6132{\scriptsize$\pm$.6132} & $\upuparrows$   & 0.5027{\scriptsize$\pm$.0521} & 0.5239{\scriptsize$\pm$.0525}& $\upuparrows$    \\
enron     & 53    & 0.3843{\scriptsize$\pm$.3843} & 0.4016{\scriptsize$\pm$.4016} & $\upuparrows$   & 0.2821{\scriptsize$\pm$.0190} & 0.3005{\scriptsize$\pm$.0238}& $\upuparrows$    \\
genbase   & 27    & 0.9843{\scriptsize$\pm$.9843} & 0.9838{\scriptsize$\pm$.9838} &           & 0.9807{\scriptsize$\pm$.0129} & 0.9802{\scriptsize$\pm$.0125}&          \\
image     & 5     & 0.5557{\scriptsize$\pm$.5557} & 0.5315{\scriptsize$\pm$.5315} & $\downdownarrows$ & 0.5197{\scriptsize$\pm$.0272} & 0.4972{\scriptsize$\pm$.0304}& $\downdownarrows$  \\
mediamill   & 101     & 0.5328{\scriptsize$\pm$.0054} & 0.5610{\scriptsize$\pm$.0052} & $\upuparrows$   & 0.4282{\scriptsize$\pm$.0052} & 0.4457{\scriptsize$\pm$.0050}& $\upuparrows$    \\
medical   & 45    & 0.6220{\scriptsize$\pm$.6220} & 0.6231{\scriptsize$\pm$.6231} &           & 0.5898{\scriptsize$\pm$.0435} & 0.5900{\scriptsize$\pm$.0460}&          \\
reuters     & 7     & 0.8624{\scriptsize$\pm$.8624} & 0.8639{\scriptsize$\pm$.8639} &           & 0.8367{\scriptsize$\pm$.0142} & 0.8382{\scriptsize$\pm$.0126}&          \\
scene     & 6     & 0.5921{\scriptsize$\pm$.5921} & 0.6105{\scriptsize$\pm$.6105} & $\upuparrows$   & 0.5739{\scriptsize$\pm$.0423} & 0.5873{\scriptsize$\pm$.0370}& $\upuparrows$    \\
slashdot    & 22    & 0.3271{\scriptsize$\pm$.3271} & 0.3248{\scriptsize$\pm$.3248} &           & 0.2843{\scriptsize$\pm$.0186} & 0.2818{\scriptsize$\pm$.0202}&          \\
yeast     & 14    & 0.5889{\scriptsize$\pm$.5889} & 0.6141{\scriptsize$\pm$.6141} & $\upuparrows$   & 0.4890{\scriptsize$\pm$.0200} & 0.5104{\scriptsize$\pm$.0200}& $\upuparrows$    \\
\hline
\vspace{0.1cm}\\
 &  & \multicolumn{2}{c}{\textsc{Hamming Loss}} & & \multicolumn{2}{c}{\textsc{Subset 0/1 Loss}} & \\
 \hline
  & $m$ & $\text{CC}_{SC}$ & $\text{NS}_{SC}$ & & $\text{CC}_{SC}$ & $\text{NS}_{SC}$ \\
 \hline 
bibtex& 159   & 0.0282{\scriptsize$\pm$.0008} & 0.0270{\scriptsize$\pm$.0006}&$\upuparrows$   & 0.9592{\scriptsize$\pm$.0080} & 0.9568{\scriptsize$\pm$.0082}& $\uparrow$     \\ 
emotions &6   & 0.2363{\scriptsize$\pm$.0268} & 0.2190{\scriptsize$\pm$.0266}&$\upuparrows$   & 0.7555{\scriptsize$\pm$.0581} & 0.7404{\scriptsize$\pm$.0652}& $\uparrow$     \\
enron & 53    & 0.0819{\scriptsize$\pm$.0023} & 0.0766{\scriptsize$\pm$.0030}&$\upuparrows$   & 0.9491{\scriptsize$\pm$.0130} & 0.9346{\scriptsize$\pm$.0156}& $\upuparrows$    \\
genbase &27   & 0.0019{\scriptsize$\pm$.0012} & 0.0019{\scriptsize$\pm$.0012}&          & 0.0332{\scriptsize$\pm$.0176} & 0.0337{\scriptsize$\pm$.0172}&          \\
image & 5     & 0.2104{\scriptsize$\pm$.0127} & 0.2199{\scriptsize$\pm$.0140}&$\downdownarrows$ & 0.5855{\scriptsize$\pm$.0270} & 0.6027{\scriptsize$\pm$.0277}& $\downdownarrows$  \\
mediamill&101 & 0.0302{\scriptsize$\pm$.0004} & 0.0291{\scriptsize$\pm$.0003}&$\upuparrows$   & 0.8750{\scriptsize$\pm$.0049} & 0.8925{\scriptsize$\pm$.0051}& $\downdownarrows$  \\
medical & 45  & 0.0210{\scriptsize$\pm$.0025} & 0.0210{\scriptsize$\pm$.0027}&          & 0.5017{\scriptsize$\pm$.0465} & 0.5037{\scriptsize$\pm$.0514}&          \\
reuters & 7   & 0.0513{\scriptsize$\pm$.0049} & 0.0506{\scriptsize$\pm$.0042}&          & 0.2403{\scriptsize$\pm$.0177} & 0.2391{\scriptsize$\pm$.0167}&          \\
scene & 6     & 0.1479{\scriptsize$\pm$.0147} & 0.1441{\scriptsize$\pm$.0130}&$\uparrow$      & 0.4802{\scriptsize$\pm$.0449} & 0.4815{\scriptsize$\pm$.0386}&          \\
slashdot& 22  & 0.0840{\scriptsize$\pm$.0026} & 0.0842{\scriptsize$\pm$.0028}&          & 0.8348{\scriptsize$\pm$.0186} & 0.8380{\scriptsize$\pm$.0201}&          \\
yeast & 14    & 0.2243{\scriptsize$\pm$.0093} & 0.2089{\scriptsize$\pm$.0097}&$\upuparrows$   & 0.8073{\scriptsize$\pm$.0230} & 0.8097{\scriptsize$\pm$.0237}&          \\
\hline
\end{tabular}
}
\label{tab:NSscvsCCsc}
\end{table}

\begin{table}[th]
\caption{The effect of subset correction in terms of statistical significance. The corresponsing loss/accuracy values can be found in Tables~\ref{tab:NSvsCC}-\ref{tab:NSscvsCCsc}. $\upuparrows$ ($\downdownarrows$) means that $\text{NS}_{SC}$ ($\text{CC}_{SC}$) is significantly better (worse) than NS (CC) at level $p<0.01$ ($\uparrow$ and $\downarrow$ at level $p<0.05$) in a paired t-test.}
\centering
\begin{tabular}{lc@{\hspace{2em}}c@{\hspace{1em}}cl@{\hspace{2em}}c@{\hspace{1em}}cl} 
%\hline
%\vspace{-1.22pt}
 &  & \multicolumn{5}{c}{NS vs. $\text{NS}_{SC}$} & \\
 \hline
  & $m$  & Hamming & Subset 0/1 & & Jaccard & $F_{1}$ & \\
 \hline 
bibtex    &159  & $\upuparrows$   & $\upuparrows$&  & $\upuparrows$ & $\upuparrows$\\
emotions  &6    &           & $\upuparrows$&  & $\upuparrows$ & $\upuparrows$\\
enron   &53   & $\upuparrows$   & $\upuparrows$&  & $\upuparrows$ & $\upuparrows$\\
genbase   &27   &           & $\upuparrows$&  &         &   \\
image   &5    & $\downdownarrows$ & $\upuparrows$&  & $\upuparrows$ & $\upuparrows$\\
mediamill &101  & $\upuparrows$   & $\upuparrows$&  &         & $\downdownarrows$\\
medical   &45   & $\upuparrows$   & $\upuparrows$&  & $\uparrow$  & $\downdownarrows$\\
reuters   &7    & $\downdownarrows$ & $\upuparrows$&  & $\upuparrows$ & $\upuparrows$\\
scene   &6    & $\downdownarrows$ & $\upuparrows$&  & $\upuparrows$ & $\upuparrows$\\
slashdot  &22   & $\upuparrows$   & $\upuparrows$&  & $\upuparrows$ &   \\
yeast   &14   & $\downdownarrows$ & $\upuparrows$&  & $\upuparrows$ & $\upuparrows$\\
\hline
\vspace{0.1cm}\\
 &  &  \multicolumn{5}{c}{CC vs. $\text{CC}_{SC}$}  & \\
 \hline
  & $m$  & Hamming & Subset 0/1 & & Jaccard & $F_{1}$ & \\
 \hline 
bibtex    &159  & $\upuparrows$   & $\upuparrows$ & & $\upuparrows$   & $\upuparrows$\\
emotions  &6    &           & $\uparrow$  & &           &   \\
enron   &53   & $\upuparrows$   & $\upuparrows$ & & $\upuparrows$   & $\upuparrows$\\
genbase   &27   &           & $\upuparrows$ & &           &   \\
image   &5    &           &         & &           &   \\
mediamill &101  & $\upuparrows$   & $\upuparrows$ & & $\upuparrows$   & $\upuparrows$\\
medical   &45   & $\upuparrows$   & $\upuparrows$ & &           & $\downdownarrows$\\
reuters   &7    & $\downdownarrows$ & $\upuparrows$ & & $\upuparrows$   & $\upuparrows$\\
scene   &6    & $\downarrow$    & $\upuparrows$ & &           & $\downdownarrows$\\
slashdot  &22   & $\upuparrows$   & $\upuparrows$ & & $\upuparrows$   &   \\
yeast   &14   &           & $\upuparrows$ & & $\upuparrows$   & $\upuparrows$\\
\hline
\end{tabular}
\label{tab:SC}
\end{table}

Looking at the comparison between CC and NS (without subset correction) as shown in Table~\ref{tab:NSvsCC}), the first thing to mention is the strong performance of NS in terms of Hamming loss (8 significant wins and 3 losses). In terms of their properties, the three data sets on which NS loses do indeed seem to be favorable for CC: Since slashdot, medical and genbase all have a rather low Hamming loss, the danger of error propagation is limited. Thus, the results are completely in agreement with our expectations. 
%line with our reasoning on the potential failure of CC.

For Jaccard and F1, the picture is not as clear. In both cases, NS wins 6 times. Again, like for Hamming loss, NS outperforms CC on data sets with many labels (bibtex, enron, mediamill) or a relatively high Hamming loss (yeast), whereas CC is better for data sets with only a few labels (image, reuters) or with high accuracy (genbase). 
%Overall, NS tends to outperform CC on data sets with many labels or low (label-wise) prediction accuracy. 

The picture for CC and NS with subset correction (denoted $\text{CC}_{SC}$ and $\text{NS}_{SC}$, respectively) is quite similar (Table~\ref{tab:NSscvsCCsc}), although the performance differences tend to decrease in absolute size. 
%Overall, NS seems to benefit more from subset correction than CC. For example, 
On subset 0/1 loss, for which the original CC performs quite strong and typically outperforms NS, the corrected version $\text{NS}_{SC}$ even achieves 3 significant wins over $\text{CC}_{SC}$.

To analyze the effect of subset correction in more detail, Table~\ref{tab:SC} provides a summary of a comparison of Table~\ref{tab:NSvsCC} and Table~\ref{tab:NSscvsCCsc}. Interestingly enough, subset correction yields improvements on almost every experiment, regardless of the performance measure, and most of these improvements are even significant. More specifically, counting the number of significant wins, subset correction appears to be most beneficial for subset 0/1 loss and least beneficial for Hamming loss. In fact, for Hamming loss, subset correction loses for data sets with only a few labels (reuters, scene, yeast and image) and a relatively high observation rate.
Comparing NS and CC, the former seems to benefit even more from subset correction than the latter, except for Hamming loss, on which NS is already strong in its basic version. In terms of subset 0/1 loss, however, significant improvements can be seen on every single data set. In light of the simplicity of the idea, these effects of subset correction are certainly striking. 

% simplicity, the authors are not aware of any work, which proposes or examines the procedure so far. 

%CC it is beneficial in almost every case to apply the subset correction procedure, whereas for NS this is also true except for Hamming loss, where the picture is not as clear.

\section{Conclusion}\label{sec:conclusions}

This paper has thrown a critical look at the classifier chains method for multi-label classification, which has been adopted quite quickly by the MLC community and is now commonly used as a baseline when it comes to comparing methods for exploiting label dependency. Notwithstanding the appeal of the method and the plausibility of its basic idea, we have argued that, at second sight, the chaining of classifiers begs an important flaw: A binary classifier that has learnt to rely on the values of previous labels in the chain might be misled when these values are replaced by possibly erroneous estimations at prediction time. The classification errors produced because of this attribute noise may subsequently be propagated or even multiplied along the entire chain. Roughly speaking, what looks as a gift at training time may turn out to become a handicap in prediction. 

Our results have shown that the problem of error propagation is highly relevant, and that it may strongly impair the performance of CC. In order to avoid this problem, the method of nested stacking proposed in this paper uses predicted instead of observed label relevances as additional attribute values in the training phase. Our experimental studies clearly confirm that, although NS does not consistently outperform CC, it seems to have advantages for those data sets on which error propagation becomes an issue, namely data sets with many labels or low (label-wise) prediction accuracy.

%First, a modification has been proposed for its avoidance, namely nested stacking and second, a more general procedure, subset correction, has been introduced and the effectiveness of both of them has been demonstrated empirically. 

There are several lines of future work. First, it is of course desirable to complement this study by meaningful theoretical results supporting our claims. Second, it would be interesting to investigate to what extent the problem of attribute noise also applies to the probabilistic variant of classifier chains introduced in \cite{EykeBayes}. Last but not least, given the interesting effects that are produced by the simple idea of subset correction, this approach seems to be worth further investigation, all the more as it is completely general and not limited to specific MLC methods such as those considered in this paper.

%\begin{acknowledgements}
%\end{acknowledgements}

%\bibliography{lit}   % name your BibTeX data base

\begin{thebibliography}{10}
\providecommand{\url}[1]{{#1}}
\providecommand{\urlprefix}{URL }
\expandafter\ifx\csname urlstyle\endcsname\relax
  \providecommand{\doi}[1]{DOI~\discretionary{}{}{}#1}\else
  \providecommand{\doi}{DOI~\discretionary{}{}{}\begingroup
  \urlstyle{rm}\Url}\fi

\bibitem{cauwenberghs2001incremental}
Cauwenberghs, G., Poggio, T.: Incremental and decremental support vector
  machine learning.
\newblock Proc.\ NIPS pp. 409--415 (2001)

\bibitem{EykeIBLR}
Cheng, W., H\"{u}llermeier, E.: {Combining instance-based learning and logistic
  regression for multilabel classification}.
\newblock Machine Learning \textbf{76}(2-3), 211--225 (2009).
\newblock \doi{10.1007/s10994-009-5127-5}.

\bibitem{Cohen_05}
Cohen, W.W.: Stacked sequential learning.
\newblock Tech. rep., DTIC Document (2005)

\bibitem{EykeBayes}
Dembczy\'{n}ski, K., Cheng, W., H\"{u}llermeier, E.: Bayes optimal multilabel
  classification via probabilistic classifier chains.
\newblock In: ICML, pp. 279--286 (2010)

\bibitem{LabelDependenceML}
Dembczy\'{n}ski, K., Waegeman, W., Cheng, W., H\"{u}llermeier, E.: On label
  dependence and loss minimization in multi-label classification.
\newblock Machine Learning \textbf{88}, 5--45 (2012)

\bibitem{Dietterich_95}
Dietterich, T.G., Bakiri, G.: Solving multiclass learning problems via
  error-correcting output codes.
\newblock Journal of Artificial Intelligence Research \textbf{2}, 263--286
  (1995)

\bibitem{doppa2}
Doppa, J.R., Fern, A., Tadepall, P.: {HC}-{S}earch: {L}earning heuristics and
  cost functions for structured prediction.
\newblock In: Proc.\ AAAI, National Conference on Artificial Intelligence
  (2012)

\bibitem{doppa}
Doppa, J.R., Fern, A., Tadepall, P.: Output space search for structured
  prediction.
\newblock In: Proc.\ ICML, International Conference on Machine Learning.
  Scotland, UK (2012)

\bibitem{WestonMulti-Label}
Elisseeff, A., Weston, J.: {A Kernel Method for Multi-Labelled Classification}.
\newblock In: ACM Conf. on Research and Develop. in Infor. Retrieval, pp.
  274--281 (2005).
\newblock
 
\bibitem{FurnkranzLabelRanking}
F\"{u}rnkranz, J., H\"{u}llermeier, E., Menc\'{\i}a, E., Brinker, K.:
  Multilabel classification via calibrated label ranking.
\newblock Machine Learning \textbf{73}, 133--153 (2008).
\newblock \doi{10.1007/s10994-008-5064-8}.

\bibitem{Fuernkranz_12}
F\"{u}rnkranz, J., Park, S.H.: Error-correcting output codes as a
  transformation from multi-class to multi-label prediction.
\newblock In: Proc.\ Discovery Science, pp. 254--267 (2012).
\newblock \doi{10.1007/978-3-642-33492-4_21}.

\bibitem{GodboleDiscriminative}
Godbole, S., Sarawagi, S.: Discriminative methods for multi-labeled
  classification.
\newblock In: Pacific-Asia Conf. on Know. Disc. and Data Mining, pp. 22--30
  (2004)

\bibitem{searn}
III, H.D., Langford, J., Marcu, D.: Search-based structured prediction.
\newblock Machine Learning \textbf{75}(3), 297--325 (2009)

\bibitem{Kajdanowicz_12}
Kajdanowicz, T., Kazienko, P.: Multi-label classification using error
  correcting output codes.
\newblock International Journal of Applied Mathematics and Computer Science
  \textbf{22}(4), 829--840 (2012)

\bibitem{liblr08}
Lin, C.J., Weng, R.C., Keerthi, S.S.: Trust region {N}ewton method for logistic
  regression.
\newblock Journal of Machine Learning Research \textbf{9}(Apr), 627--650 (2008)

\bibitem{elenaAID}
Monta{\~n}{\'e}s, E., Quevedo, J.R., del Coz, J.J.: Aggregating independent and
  dependent models to learn multi-label classifiers.
\newblock In: Proc.\ ECML/PKDD (2011)

\bibitem{nguyen07}
Nguyen, N., Guo, Y.: Comparisons of sequence labeling algorithms and
  extensions.
\newblock In: Proc.\ ICML, International Conference on Machine Learning (2007)

\bibitem{ReadPrunedSets}
Read, J., Pfahringer, B., Holmes, G.: Multi-label classification using
  ensembles of pruned sets.
\newblock In: IEEE Int. Conf. on Data Mining, pp. 995--1000. IEEE (2008).
\newblock \doi{10.1109/ICDM.2008.74}.

\bibitem{ClassifierChainsML}
Read, J., Pfahringer, B., Holmes, G., Frank, E.: Classifier chains for
  multi-label classification.
\newblock Machine Learning \textbf{85}(3), 333--359 (2011)

\bibitem{Ross_10}
Ross, S., Bagnell, D.: Efficient reductions for imitation learning.
\newblock In: International Conference on Artificial Intelligence and
  Statistics, pp. 661--668 (2010)

\bibitem{SchapireBoostexter}
Schapire, R.E., Singer, Y.: Boostexter: A boosting-based system for text
  categorization.
\newblock In: Machine Learning, pp. 135--168 (2000)

\bibitem{Senge_GfKl_2012}
Senge, R., del Coz, J.J., H\"{u}llermeier, E.: On the problem of error
  propagation in classifier chains for multi-label classification.
\newblock In: Conference of the German Classification Society (2012)

\bibitem{TsoumakasMedidasError}
Tsoumakas, G., Katakis, I., Vlahavas, I.: Mining multi-label data.
\newblock In: Data Mining and Knowledge Discovery Handbook, pp. 667--685 (2010)

\bibitem{TsoumakasRAKEL}
Tsoumakas, G., Vlahavas, I.: {Random k-Labelsets: An Ensemble Method for
  Multilabel Classification}.
\newblock In: Proc.\ ECML/PKDD, LNCS, pp. 406--417. Springer (2007).
\newblock \doi{10.1007/978-3-540-74958-5\_38}.

\bibitem{WolpertStacked}
Wolpert, D.H.: Stacked generalization.
\newblock Neural Networks \textbf{5}, 214--259 (1992)

\bibitem{ZhangML-NN}
Zhang, M.L., Zhou, Z.H.: Multilabel neural networks with applications to
  functional genomics and text categorization.
\newblock IEEE Trans. on Knowl. and Data Eng. \textbf{18}, 1338--1351 (2006).
\newblock \doi{http://dx.doi.org/10.1109/TKDE.2006.162}.

\bibitem{ZhuNoise}
Zhu, X., Wu, X.: Class noise vs. attribute noise: a quantitative study of their
  impacts.
\newblock Artificial Intelligence Review \textbf{22}(3), 177--210 (2004).
\newblock \doi{10.1007/s10462-004-0751-8}

\end{thebibliography}

\end{document}